\pdfoutput=1
\documentclass{article}

\usepackage[a4paper, total={6in, 8in}]{geometry}
\usepackage[utf8]{inputenc}
\usepackage[T1]{fontenc}
\usepackage{authblk}
\usepackage{graphicx}
\usepackage{url}
\usepackage{lipsum}
\usepackage[normalem]{ulem}
\useunder{\uline}{\ul}{}
\title{On Clustering Categories of Categorical Predictors in Generalized Linear Models}

\usepackage{textgreek}
\usepackage[round]{natbib}
 \usepackage{setspace}
\usepackage[ruled,linesnumbered]{algorithm2e}
\usepackage{amssymb}
\usepackage{caption}
\include{pythonlisting}
\usepackage{lscape}
\usepackage{amsmath}
\usepackage{enumitem}
\usepackage[english]{babel}
\usepackage{xcolor}
\usepackage{algorithmic}\usepackage{multirow}
\usepackage{hyperref}

\usepackage[linesnumbered]{algorithm2e}
\date{} 
\providecommand{\keywords}[1]
{
  \small	
  \textbf{\textit{Keywords---}} #1
}

\author[1]{Emilio Carrizosa}
\affil[1]{Instituto de Matemáticas de la Universidad de Sevilla, Sevilla, Spain\newline
\tt{ecarrizosa@us.es}}
\author[2]{Marcela Galvis Restrepo}
\author[2]{Dolores Romero Morales}
\affil[2]{Copenhagen Business School, Frederiksberg, Denmark \newline
        \tt{\{mgr.eco,drm.eco\}@cbs.dk}}
\begin{document}
\graphicspath{ {images/} }
\doublespacing
\begin{titlepage}
\centering
{\bfseries\Huge On Clustering Categories of Categorical Predictors in Generalized Linear Models \par}
\vspace{1cm}
{\bfseries \Large Emilio Carrizosa, Marcela Galvis Restrepo and Dolores Romero Morales \par}
\vspace{1cm}
{\Large Journal article (Accepted manuscript*) \par}
\vspace{1cm}
{ \textbf{Please cite this article as}: Carrizosa, E., Galvis Restrepo, M., and Romero Morales, D. (2021). On clustering categories of categorical predictors in generalized linear models. Expert Systems with Applications. https://doi.org/10.1016/j.eswa.2021.115245 \par}
\vspace{1cm}
{\Large DOI: \url{https://doi.org/10.1016/j.eswa.2021.115245} \par}
\vspace{1cm}
{\Large Available online June 03 2021 \par}
\vspace{2cm}
{© 2021. This manuscript version is made available under the CC-BY-NC-ND 4.0 license \url{http://creativecommons.org/licenses/by-nc-nd/4.0/}}
\end{titlepage}
\maketitle

\begin{abstract}
We propose a method to reduce the complexity of Generalized Linear Models in the presence of categorical predictors. The traditional one-hot encoding, where each category is represented by a dummy variable, can be wasteful, difficult to interpret, and prone to overfitting, especially when dealing with high-cardinality categorical predictors. This paper addresses these challenges by finding a reduced representation of the categorical predictors by clustering their categories. This is done through a numerical method which aims to preserve (or even, improve) accuracy, while reducing the number of  coefficients to be estimated for the categorical predictors. Thanks to its design, we are able to derive a proximity measure between categories of a categorical predictor that can be easily visualized. We illustrate the performance of our approach in real-world classification and count-data datasets where we see that clustering the categorical predictors reduces complexity substantially without harming accuracy. 
\end{abstract}


\section{Introduction}

Categorical predictors are increasingly present in classification and regression applications. In many social science applications with individual-level data, categorical predictors representing group-membership, like individuals with the same country of origin, attending the same school or being treated at the same hospital, might be of interest \citep{bonhomme2015grouped,johannemann2019sufficient}. Categorical predictors can also depict complex behavioural data related to, e.g., people's demographics and preferences  \citep{moeyersoms2016explaining}. Finally, high-cardinality categorical predictors associated with words arise also in Natural Language Processing \citep{Mikolov2013EfficientEO,cerda2018similarity}. No matter the application, before using categorical predictors in any linear model, it is necessary to transform them into real-valued vectors, which is usually done by representing each category by a dummy and leaving one out for contrast, an approach usually referred as the \textit{one-hot dummy encoding}. 

The main downside when using the \textit{one-hot dummy encoding} is that to understand the effect of each categorical predictor on the linear model requires the interpretation of the coefficients of the corresponding collection of dummy variables. In the presence of high-cardinality categorical predictors, this can be a cumbersome task. To see this, consider the \textit{Adult} dataset which we use in our experimental section. For one of the $11$ categorical predictors in the dataset, namely \textit{Country}, we have $41$ categories. This means that to explain the effect of  \textit{Country} in the model requires the interpretation of $40$ coefficients, each of them with its own magnitude and sign. It would have been much easier to interpret one single coefficient, instead, as it is the case for binary predictors. There are other potential negative consequences of the one-hot dummy encoding in the presence of many categorical predictors and/or many categories, such as the risk of overfitting or having estimates of the coefficients with high uncertainty \citep{leblanc1998monotone}. To address the challenges above, we propose to cluster the categories of categorical predictors \citep{carrizosa2017clustering}, finding thus a reduced representation of the categorical predictors that requires fewer dummy variables, and thus coefficients.

In this paper, we propose a methodology to cluster the categories of categorical predictors in Generalized Linear Models (GLM). The goal is to split the categories of each categorical predictor into a number of clusters, such that categories in the same cluster have a similar impact in the model and thus can be grouped together. This clustering of the categories yields to a reduced representation of the categorical predictor in which we have as many dummy variables as clusters. Our approach achieves a more compact representation of the categorical predictors and therefore reduces the number of coefficients to estimate and interpret. To illustrate this, let us go back to our example from the \textit{Adult} dataset. The predictor \textit{Type of job} includes categories \textit{State-gov, Federal-gov, Local-gov, Self-emp, Private, Missing}. If \textit{State-gov} and \textit{Local-gov} are clustered together and the rest are in another cluster, the $5$ dummy variables representation for \textit{Type of job} would be reduced to just one dummy variable.   

Our methodology is an iterative process, in which we cluster a categorical predictor in each iteration. We propose a numerical method that in each iteration chooses the predictor to cluster and its best clustering. These decisions are made guided by out-of-sample accuracy of GLM in which the categorical predictors clustered in previous iterations are represented in their reduced form, an additional predictor is clustered, and the remaining ones stay as originally. Once all categorical predictors have been clustered, we train the clustered GLM. This process is repeated a number of times to obtain a series of clustered models, and the one with the best out-of-sample accuracy is chosen. This collection of clustered models offers a way to measure and visualize the proximity between categories of a predictor as the percentage of these clustered GLM where the categories are together in the same cluster.

In summary, our methodology has the following advantages. First, having less coefficients to interpret gives a less complex model which is a step towards enhancing the interpretability of the GLM with categorical predictors. Second, the choice of reduced representation of categorical predictors is guided by out-of-sample accuracy, with the aim that the clustered GLM preserves (or even improves) the accuracy of the original model, as illustrated in our numerical experiments. Third, we have a data driven approach to measure and visualize the similarity between categories based on a large collection of clustered models. Fourth, given that we are clustering together categories that have a similar impact in the model, we have more observations to estimate each coefficient in the clustered GLM, which ensures lower standard errors. Fifth, our approach can also be applied to regularized linear models, i.e., instead of using GLM as the base model we could have used, e.g., a LASSO \citep{van2008high,hastie2015statistical} or a Group LASSO \citep{meier2008group,detmer2020note}.

The remainder of this article is structured as follows. The next section introduces the algorithm to cluster the categories of categorical predictors in GLM. Section \ref{sec:numerical} illustrates the performance of our method for a collection real-world datasets, in terms of accuracy and complexity, as well as the proximity measure between categories. Finally, conclusions and future research are collected in Section \ref{sec:conclusions.}

\section{Methodology}
\label{sec:methodology}
 
In this section we present our approach to cluster the categories of categorical predictors in Generalized Linear Models (GLM). We first introduce the notation for the GLM with categorical predictors. We then introduce the algorithm for clustering the categories, yielding a representation of each categorical predictor $j$ with $K'$ dummy variables. With this, we build the so-called clustered GLM in which each categorical predictor is modeled using its reduced representation. We end the section discussing how to get insights into the relationship between categories of a categorical predictor using a measure of proximity that stems from our numerical method.

We are given a training sample of size $N$. We have $J$ categorical as well as $P$ continuous predictors. Categorical predictor $j$ has $K_j$ categories,  $j=1,\ldots,J$. In the GLM using the traditional one-hot encoding, a categorical predictor $j$ with $K_j$ categories is represented by $K_j-1$ dummy variables, one for each category, leaving one out for contrast. We denote by $\mathbf{d}$ the vector of dummy variables associated with the categorical predictors, while $\mathbf{x}$ denotes the vector of continuous predictors. Consider a GLM where the outcome $y$ is related to $\mathbf{d}$ and $\mathbf{x}$ through a link function ~$G$, namely, 

\begin{equation}
\label{eq:GLMmodel}
{\mathbb{E} \, [y | \mathbf{d},\mathbf{x} ]}=G(\beta_0 + (\boldsymbol\beta)^T \mathbf{d}  + (\boldsymbol{\tilde{\beta}})^T \mathbf{x}),
\end{equation}
where $\beta_0$ is the intercept, $\boldsymbol\beta$ is the set of model parameters for the dummy variables and $\boldsymbol{ \tilde{\beta}}$ the one for the continuous predictors. For a binary response variable  $y \in \{0,1\}$, a natural choice of link the function $G$ is the Logit, while for a count response variable $y \in \{0,1,2,\ldots\}$, it would be the Log Link function. Both link functions will be illustrated in Section \ref{sec:numerical}, but our approach can handle any other link function.

We will now explain how the clustering of categories for a given categorical predictor will be performed. We will assume that the categories are ordered. For instance, for ordinal categorical predictors like the level of \textit{Education} from the \textit{Adult} dataset in Section \ref{sec:numerical}, we could take the \textit{natural} order of the predictor as the order of the categories (see Table \ref{Table1}). For non-ordinal categorical predictors we could use the coefficients from the GLM with one-hot encoding to order the categories.

\begin{table}[]
\centering
\caption{{Feasible clusterings for the predictor \textit{Education} in the \textit{Adult} dataset with $K' = 2$}}
\begin{tabular}{cccccccccccccccc}
\hline
 & \multicolumn{15}{c}{Feasible clusterings} \\ \hline
\textit{1st-4th} & 1 & 1 & 1 & 1 & 1 & 1 & 1 & 1 & 1 & 1 & 1 & 1 & 1 & 1 & 1 \\
\textit{5th-6th} & 0 & 1 & 1 & 1 & 1 & 1 & 1 & 1 & 1 & 1 & 1 & 1 & 1 & 1 & 1 \\
\textit{7th-8th} & 0 & 0 & 1 & 1 & 1 & 1 & 1 & 1 & 1 & 1 & 1 & 1 & 1 & 1 & 1 \\
\textit{9th} & 0 & 0 & 0 & 1 & 1 & 1 & 1 & 1 & 1 & 1 & 1 & 1 & 1 & 1 & 1 \\
\textit{10th} & 0 & 0 & 0 & 0 & 1 & 1 & 1 & 1 & 1 & 1 & 1 & 1 & 1 & 1 & 1 \\
\textit{11th} & 0 & 0 & 0 & 0 & 0 & 1 & 1 & 1 & 1 & 1 & 1 & 1 & 1 & 1 & 1 \\
\textit{12th} & 0 & 0 & 0 & 0 & 0 & 0 & 1 & 1 & 1 & 1 & 1 & 1 & 1 & 1 & 1 \\
\textit{HS-grad} & 0 & 0 & 0 & 0 & 0 & 0 & 0 & 1 & 1 & 1 & 1 & 1 & 1 & 1 & 1 \\
\textit{Some-college} & 0 & 0 & 0 & 0 & 0 & 0 & 0 & 0 & 1 & 1 & 1 & 1 & 1 & 1 & 1 \\
\textit{Assoc-voc} & 0 & 0 & 0 & 0 & 0 & 0 & 0 & 0 & 0 & 1 & 1 & 1 & 1 & 1 & 1 \\
\textit{Assoc-acdm} & 0 & 0 & 0 & 0 & 0 & 0 & 0 & 0 & 0 & 0 & 1 & 1 & 1 & 1 & 1 \\
\textit{Prof-school} & 0 & 0 & 0 & 0 & 0 & 0 & 0 & 0 & 0 & 0 & 0 & 1 & 1 & 1 & 1 \\
\textit{Bachelors} & 0 & 0 & 0 & 0 & 0 & 0 & 0 & 0 & 0 & 0 & 0 & 0 & 1 & 1 & 1 \\
\textit{Masters} & 0 & 0 & 0 & 0 & 0 & 0 & 0 & 0 & 0 & 0 & 0 & 0 & 0 & 1 & 1 \\
\textit{Doctorate} & 0 & 0 & 0 & 0 & 0 & 0 & 0 & 0 & 0 & 0 & 0 & 0 & 0 & 0 & 1 \\ \hline
\end{tabular}
\label{Table1}
\end{table}

Now let us define the concept of feasible clustering for categorical predictor $j$: we will say that a clustering of the categories of $j$ into $K'$ clusters is feasible, if each of the clusters consists of consecutive categories. For the \textit{Education} level predictor, with $K_j=15$, there are 15 feasible clusterings with $K'=2$ clusters (see Table \ref{Table1}). The first clustering corresponds to having the lowest level of education (\textit{1st-4th}) in one group, and the remaining levels in another one. The second clustering corresponds to having the first and second level of education (\textit{1st-4th}, \textit{5th-6th}) in one group and the remaining levels in another one. We successively move to include higher levels of education one by one in the first cluster, until we reach the last level, where all the categories are together in the first cluster and the second cluster is empty. This last clustering is equivalent to removing predictor $j$ entirely from the model.

Once we have a feasible clustering, we can obtain the reduced representation of the categorical predictor with $K'$ dummy variables, where we have a dummy variable for each of the $K'$ clusters. For $K'=2$, the reduced representation consists of one single dummy variable indicating whether the category belongs to the first cluster or not. This is illustrated in Table \ref{Table1} for the categorical predictor \textit{Education.}

We next argue the necessity to use a randomized numerical method to decide which feasible clustering will be used in our clustered GLM for each categorical predictor. We might decide that the best choice is that one with which the GLM achieves the highest out-of-sample accuracy. However, clusterings that differ little may yield a very similar out-of-sample accuracy. In the presence of multiple predictors to be clustered, it may desirable to choose a clustering with a good out-of-sample accuracy, but not necessarily the best one. Therefore, we design a Greedy Randomized Adaptive Search Procedure (GRASP) that chooses randomly between the feasible clusterings with the highest out-of-sample accuracies, see Figure \ref{fig:algo.}

GRASP is a class of numerical methods that have been successfully applied to a number of optimization problems \citep{resende2016optimization}. GRASP deals with optimization problems in which a collection of decisions needs to be made, and for each decision one has a number of feasible actions. The decisions have an associated payoff, and the goal is to maximize the payoff. GRASP makes in each step a new decision in a random fashion, choosing from the top $h\%$ payoffs. GRASP is repeated $m$ times, and the solution with the best payoff across the $m$ iterations is returned. In our case, GRASP needs to decide in each step which categorical predictor to cluster, the feasible actions are the feasible clusterings for the corresponding predictor, while the payoff is the out-of-sample accuracy of the GLM where the categorical predictor at hand is represented by the dummy variables associated with the feasible clustering. Once the feasible clustering is chosen, the predictor is clustered for the steps to come. In this way, after all categorical predictors have been clustered into $K’$ clusters, we can build the corresponding clustered GLM. These steps are repeated $m$ times, and GRASP returns the best clustered GLM across the $m$ iterations performed in terms of out-of-sample accuracy.

\begin{figure}
\centering
 
\caption{ Pseudocode for the GRASP algorithm}
\label{fig:algo}
  \begin{algorithm}[H] 
\SetAlgoLined
  \textbf{Initialization:} Let $\mathcal{L} \subseteq \{1,2,\ldots,J\}$ be the set of categorical predictors to be clustered. Let $\mathbf{d}$ be the vector of dummy variables in the one-hot-encoding and $\mathbf{x}$ the vector of continuous predictors. Let $K'<K_j-1$, for all $j \in \mathcal{L}$\; 
\For{$i \in \{1,\ldots,m\}$}{
Set $\mathcal{L'} =  \mathcal{L}$\;
\While{$\mathcal{L'} \neq  \emptyset $}{
\For{$j \in \mathcal{L'}$}{
\For{each feasible clustering of predictor $j$ with $K'$ clusters}{
	Estimate the GLM in \eqref{eq:GLMmodel} where predictors in $(\mathcal{L} \setminus \mathcal{L'}) \cup \{j\}$ are clustered\;
	Calculate its out-of-sample accuracy\;} 
	\textbf{Return: }$\mathcal{V}_j$, the set of out-of-sample accuracies\;}
    Let $\mathcal{V}=\cup_{j \in \mathcal{L'}} \mathcal{V}_j$ be the merged set of out-of-sample accuracies\;
    Sort $\mathcal{V}$ from max to min\;
    Choose randomly one accuracy from the top $h\%$ ones in $\mathcal{V}$. This accuracy is linked to $s \in \mathcal{L'}$ and a reduced representation with $K'$ dummy variables\;
    Replace the $K_s-1$ dummy variables of $s$ by the new $K'$ dummy variables and eliminate $s$ from $\mathcal{L'}$\;
    } 
    \textbf{Return: } $\mbox{GLM}^C_i$, the GLM in \eqref{eq:GLMmodel} where predictors in $\mathcal{L}$ are clustered\;
     }
\textbf{Return: } The clustered GLM, $\mbox{GLM}^C_i$ with the highest out-of-sample accuracy
 \end{algorithm} 
\end{figure}

We will now describe how from the $m$ iterations of GRASP, we do not only obtain a less complex model, which hopefully has a similar or even better accuracy, but we can also derive a proximity measure for the categories of each categorical predictor. For a given categorical predictor $j$, we can exploit the $m$ clustered GLMs built across all iterations to measure the proximity between categories $c$ and $d$. We define the proximity between categories $c$ and $d$ as the percentage of these clustered GLMs where $c$ and $d$ are together in the same cluster. The higher this proximity, the closer is their impact in the GLMs, and the more plausible to cluster them together. In the next section, we will visualize these proximities to better understand the underlying structure between the categories. 

We end noting that the methodology proposed in this section can be easily extended to discretize continuous predictors \citep{CarrizosaIJOC10} or to have as base model a regularized linear model, such as the LASSO or the Group LASSO.

\section{Experimental results}
\label{sec:numerical}

In this section we illustrate how our methodology performs in several real-world datasets. Our aim is to empirically analyse  the effect of clustering categories in a baseline classification or regression method in terms of accuracy and relative complexity. As baseline procedure for classification we have chosen logistic regression. We cluster all our categorical predictors into $K'=2$ clusters. Accuracy is measured by correct classification rate and relative complexity is the number of estimated coefficients for the categorical predictors compared to the number of estimated coefficients in the original model. In other words, the relative complexity of the clustered model is $\frac{J}{\sum_{j=1}^JK_j-J} \cdot 100\%$. Accuracy estimates are obtained as follows:  the data set is split into training sample  ($70\%$) and testing sample ($30\%$). The model is built in the training sample, and we report its accuracy on the testing sample. The process is repeated ten times and we report as estimate the average out-of-sample accuracy. We have also empirically analysed the effect of clustering categories in a baseline regression method, namely Poisson regression for count data, in terms of accuracy (measured as Root Mean Square Error) and relative complexity.

Our method uses a GRASP algorithm, as described in Figure 1. In our experiments the GRASP parameters are set to $m=100$ iterations, and selection is done out of the top $3$ accuracies for categorical predictors with more than $5$ categories and out of the top $2$ otherwise. We coded our method in R and conducted our experiments in a Workstation with an Intel\textsuperscript{\textregistered} Core\textsuperscript{TM} i5-4460 processor with 8 Gb of RAM.

The rest of the section is organized as follows. The datasets are described in Section \ref{sec:datasets}, performance in terms of accuracy and relative complexity is discussed in Section \ref{sec:performance}, and proximity graphs are shown in Section \ref{sec:proximity}. 

\subsection{Datasets
\label{sec:datasets}}

We use eight real-world classification datasets to illustrate the method, which are available in the UCI Machine Learning Repository \citep{Dua:2019}. We also use one count-data dataset, which is available in \cite{deb1997demand}. The datasets are described in Table \ref{Table2}, in the first two columns we report the name and total number of records in the dataset ($N$). In column three to seven we report the class split in percentage, the number of categorical ($J$) and continuous predictors ($P$), the total number of categories ($\sum_{j=1}^JK_j$) across all categorical predictors, and the number of categories for each categorical predictor $K_j$, respectively. Please note that \textit{Coil-2000}, \textit{Car Evaluation}, \textit{Solar} and \textit{Nursery} are converted into two-class problems using the majority class against the rest.

\begin{table}[]
\centering
\caption{{Description of the classification datasets (first eight ones) and regression dataset (last one)}}\begin{tabular}{ccccccc}\hline
\multicolumn{1}{c}{Name} & \multicolumn{1}{c}{$N$} & \multicolumn{1}{c}{Class-split} & \multicolumn{1}{c}{$J$} & \multicolumn{1}{c}{$P$} & \multicolumn{1}{c}{$\sum_{j=1}^JK_j$} & \multicolumn{1}{c}{$K_j$}      \\\hline
\textit{Solar}             & 1066                    & 83/17                           & 5                     & 5                      & 23                                & 7,6,4,3,3                           \\
\textit{Coil-2000}         & 5822                    & 94/6                            & 5                     & 80                     & 77                                & 41,6,10,10,10                       \\

\textit{Nursery}           & 12960                   & 33/66                           & 7                     & 1                      & 25                                & 3,5,4,4,3,3,3                       \\
\textit{Mushrooms}         & 8124                    & 48/52                           & 17                    & 4                      & 111                               & 6,4,10,9,4,3,12,4,4,9,9,4,3,8,9,6,7 \\
\textit{Bank marketing}    & 4119                    & 89/11                           & 6                     & 14                     & 42                                & 12,4,8,10,5,3                       \\
\textit{Car evaluation}    & 1728                    & 30/70                           & 6                     & 0                      & 21                                & 4,4,4,3,3,3                         \\
\textit{Adult}             & 32561                   & 24/76                           & 11                    & 3                      & 117                               & 5,8,5,16,5,7,14,6,5,5,41            \\
\textit{German}            & 1000                    & 30/70                           & 11                    & 9                      & 52                                & 4,5,11,5,5,5,3,4,3,3,4    \\
\textit{DebTrivedi} & 4406                        &   ---    & 5 & 5   & 25                                                    & 3,5,7,4,6     \\\hline
\end{tabular}
\label{Table2}
\end{table}
\subsection{Performance of the original and clustered models}
\label{sec:performance}
In this section we discuss the accuracy and the relative complexity. We start with the classification task. We will say that the original and the clustered models give comparable results in terms of accuracy if the difference is below 1 percentage point (p.p.). Table \ref{Table3} reports the mean testing accuracy across the ten reshuffles for accuracy as well as the relative complexity of clustered model with respect to the original one. We underline the results where the original and the clustered models give comparable accuracy.

Let us start with accuracy. For six datasets (\textit{Coil-2000, Nursery, Mushrooms, Bank marketing, Adult, German}) accuracy of the clustered model is comparable to that of the original one. For two datasets (\textit{Solar} and \textit{Car evaluation}), the original model outperforms the clustered model but only by 1.22 p.p.\ and 2.11 p.p.\ respectively. In the datasets where our model performs comparably to the original model, there is a considerable number of categories and categorical predictors. In the two datasets where the original model outperforms our method, we have categorical predictors with too few categories so it seems that by clustering them into $K'=2$ we are losing some information.

Now let us focus on the relative complexity. For \textit{Bank marketing} the number of estimated coefficients is just $5.13\%$ of the ones we would have to estimate with the original model, $10.20\%$ for \textit{Coil-2000} and $18.64\%$ for \textit{Adult}. These datasets have a considerable number of categories and/or categorical predictors and hence they benefit more from our method. On the contrary, for datasets with few categories and/or categorical predictors such as \textit{Car evaluation}, \textit{Nursery} and \textit{German}, the benefit in clustering the categories is lower although still considerable. To conclude, in our experiments we see that clustering the categorical predictors reduces complexity substantially without harming accuracy.

\begin{table}[]
\centering
\caption{Out-of-sample results for the original and the clustered model - Logistic Regression}
\begin{tabular}{cccc}\hline
\multicolumn{1}{c}{\multirow{2}{*}{Name}} & \multicolumn{2}{c}{Accuracy ($\%$)} & \multicolumn{1}{c}{\multirow{2}{*}{Relative Complexity   ($\%$)}} \\
\multicolumn{1}{c}{}                      & Original         & Clustered        & \multicolumn{1}{c}{}                                                  \\ \hline
\textit{Solar}                            & 83.89            & 82.66            & 29.41                                                                 \\
\textit{Coil-2000}                        & {\ul 93.32}      & {\ul 93.41}      & 10.20                                                                 \\
\textit{Nursery}                          & {\ul 100.00}     & {\ul 100.00}     & 38.89                                                                 \\
\textit{Mushrooms}                        & {\ul 100.00}     & {\ul 99.85}      & 21.25                                                                 \\
\textit{Bank   marketing}                 & {\ul 91.17}      & {\ul 91.38}      & 5.13                                                                  \\
\textit{Car evaluation}                   & 95.27            & 93.16            & 40.00                                                                 \\
\textit{Adult}                            & {\ul 85.09}      & {\ul 84.55}      & 18.64                                                                 \\
\textit{German}                           & {\ul 74.72}      & {\ul 74.31}      & 30.56                                                                
\\\hline                                                              
\end{tabular}
\label{Table3}
\end{table}

Finally, our method can be used for other link functions for the GLM. We show how it performs for a count data example. The baseline method will be Poisson regression. In this case, we measure accuracy by the Root Mean Square Error (RMSE), the squared root of the mean of the squares of the differences between observed and predicted values. We see there is a very small worsening of the RMSE from 4.73 to 4.86, while the number of estimated coefficients is  $41.54\%$ of the ones we would have to estimate with the original model. Hence, as in the previous results for classification, for the \textit{DebTrivedi} dataset using Poisson regression, we achieve a lower level of complexity with a slight increase in error given that the number of categories/categorical predictors is small.

\begin{table}[]
\centering
\caption{Out-of-sample results for the original and the clustered model - Poisson Regression}
\begin{tabular}{cccc}\hline
\multicolumn{1}{c}{\multirow{2}{*}{Name}} & \multicolumn{2}{c}{RMSE} & \multicolumn{1}{c}{\multirow{2}{*}{Relative Complexity   ($\%$)}} \\
\multicolumn{1}{c}{}                      & Original         & Clustered        & \multicolumn{1}{c}{}                                                  \\ \hline
\textit{DebTrivedi}                           & 4.73 & 4.86  & 41.54                                                              

\\\hline                                                              
\end{tabular}
\label{Table4}
\end{table}

\subsection{Proximity graphs}
\label{sec:proximity}

In this section we illustrate the results of the proximity measure between categories of a categorical predictor presented in Section 2. The GRASP algorithm is repeated $m$ times, out these $m$ repeats we take only the one that gives the highest accuracy and that will be our final clutered GLM. Here we dig deeper into the information from all iterations to recover valuable insights from the data, which reflects the underlying structure of the categories and their relationship in terms of proximity. We show the proximity graphs for the predictors \textit{Education}, \textit{Occupation}, and \textit{Type of Employer} from the \textit{Adult} dataset. Recall that we measure proximity as the percentage of the $m$ clustered GLMs where two categories are together in the same cluster. In the graphs, each category is a node and the thickness of the edges represents the proximity between categories. 

\begin{figure}[]
\caption{Proximity Graph for the predictor \textit{Education} in the \textit{Adult} dataset}
\centering
\includegraphics[width=15cm, height=10cm]{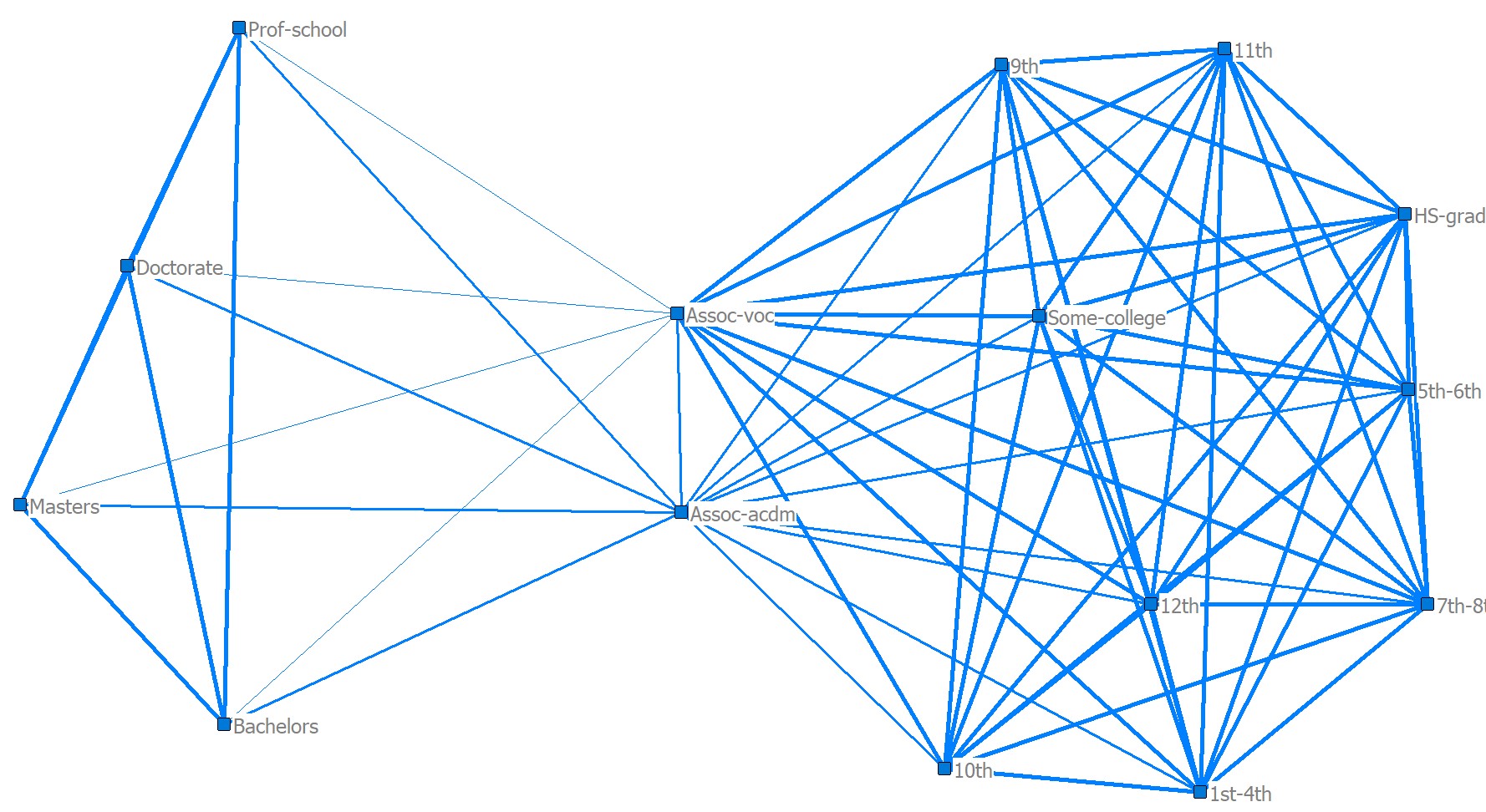}
\label{fig2}
\end{figure}

Figure \ref{fig2} shows the proximity graph for the predictor \textit{Education} in the \textit{Adult} dataset. Recall from Section 2, that this predictor is ordinal, so the categories are sorted according to its natural order: lower levels of education first and higher levels last. It is interesting to see that the cut between the two levels of education established by GRASP is well defined, with categories that represent finished university education and beyond on one cluster (left side of the graph), and those that represent unfinished and no university at all in the other (right side of the graph). In the middle we find two categories (\textit{Associate vocational} and \textit{Associate-acdm}), which are deemed as tertiary education in international classifications, but have proportionally more edges with the right hand side. 

\begin{figure}[]
\caption{Proximity Graph for the predictor \textit{Occupation} in the \textit{Adult} dataset}
\centering
\includegraphics[width=15cm, height=10cm]{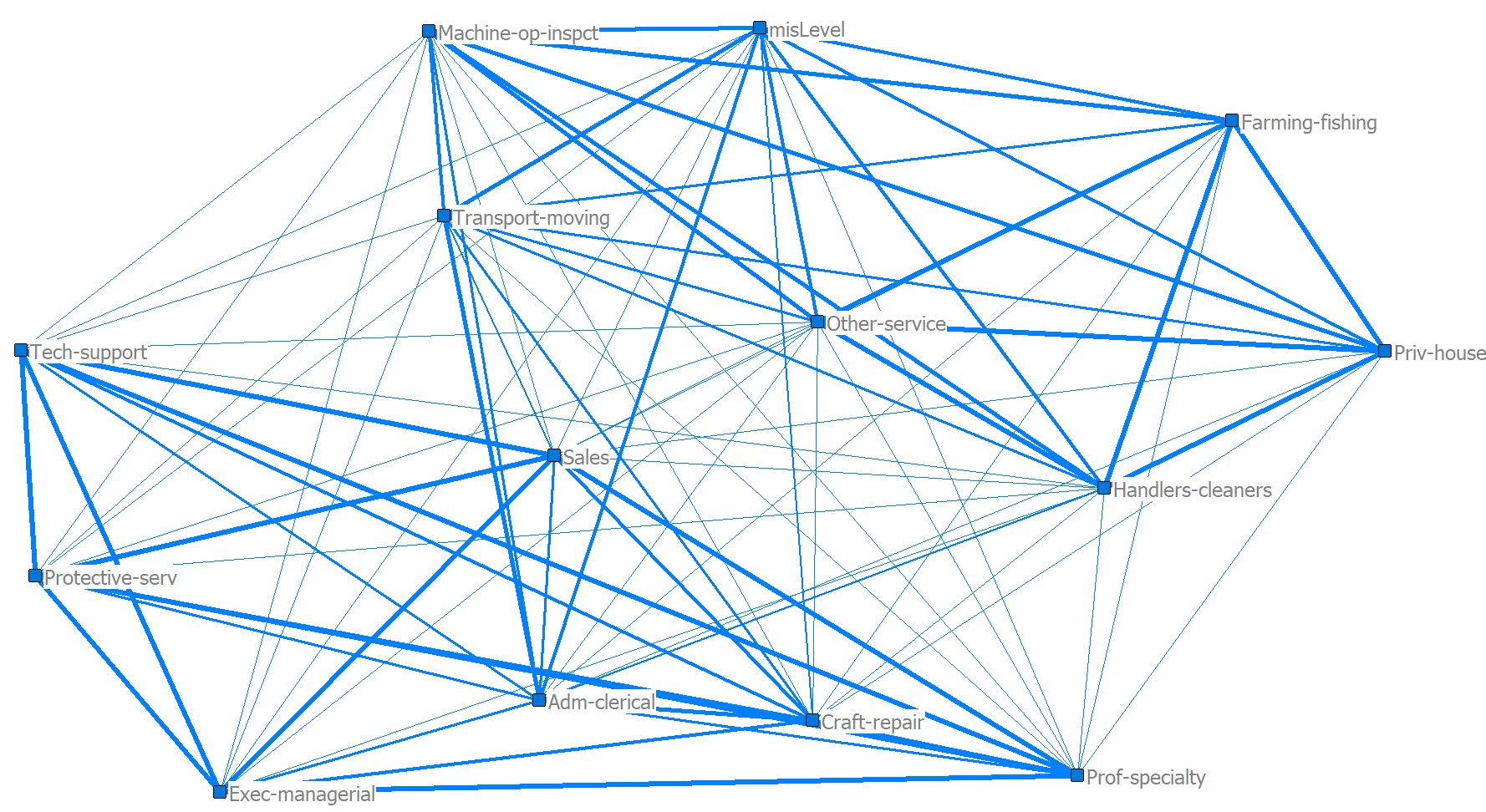}
\label{fig3}
\end{figure}

Figure \ref{fig3} shows the predictor \textit{Occupation}, for which the categories were ordered according to the coefficients of the Logistic Regression with one-hot encoding. For this predictor we can see more edges between categories, but in general the cut is established between lower paying occupations on the upper part of the graph (\textit{Farming-fishing, Priv-housing-serv, Handlers-cleaners, Transport-moving, Machine-op-insp}) and higher-paying occupations (\textit{Exec-managerial, Protective-serv, Tech-support, Sales, Craft-repair, Prof.specialty}). Some occupations like \textit{Admn-clerical} and \textit{Transport-moving} have a similar proximity to some of the lower-paying occupations as well as the high-paying occupations. 

\begin{figure}[]
\caption{Proximity Graph for the predictor \textit{Type of employer} in the \textit{Adult} dataset}
\centering
\includegraphics[width=15cm, height=10cm]{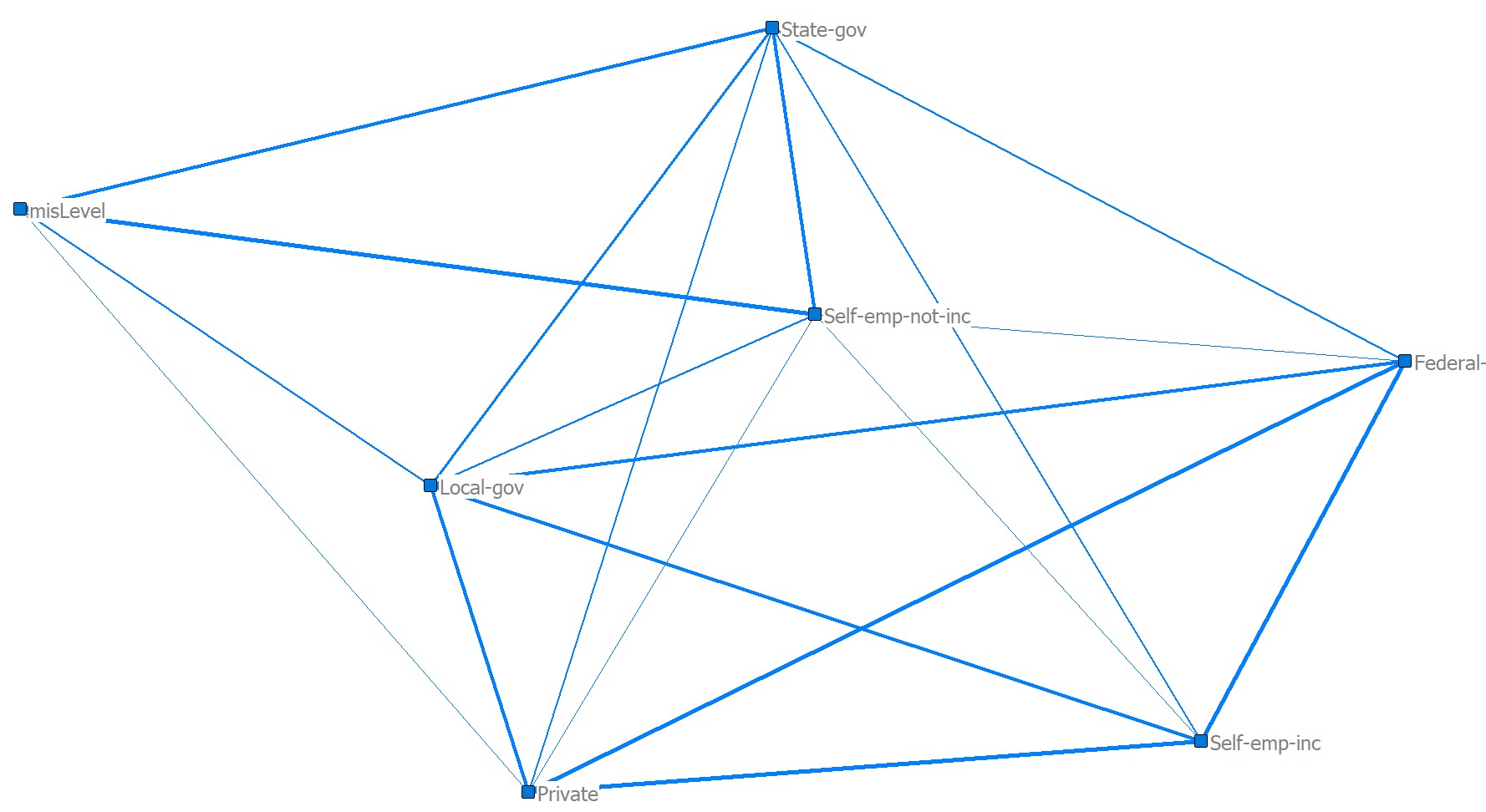}
\label{fig4}
\end{figure}

In Figure \ref{fig4} we can see the predictor \textit{Type of employer}. The order for the categories also comes from the coefficients of the Logistic Regression with one-hot encoding. The cut is less clear in terms of establishing ``high" or ``low" paying types of employer, but we can see that categories \textit{Federal, Self employed, Private and Local Government} are closer to each other, and \textit{State government} and \textit{Self-empl-not-inc} are closer with \textit{misLevel}.


\section{Conclusions}
\label{sec:conclusions}

In this paper we developed a numerical method to reduce the complexity of Generalized Linear Models in the presence of categorical predictors, by clustering their categories into $K’$ clusters. Our approach has two main advantages. First, the Clustered GLM has less coefficients to be estimated, and therefore it is easier to interpret the impact of each categorical predictor, especially when $K’=2$ and each categorical predictor is represented by one dummy variable. Second, after clustering categories, the number of observations for each cluster is higher than for the original categories, allowing for a better estimation of the coefficients. Our numerical results illustrate that it is possible to cluster categorical predictors, with a corresponding reduction in complexity, without compromising accuracy.

In terms of future research, it might be interesting to explore interactions \citep{carrizosaEJOR11} between categorical predictors which in the case of high-cardinality yields a highly combinatorial problem that needs attention in the future. In addition, other definitions of feasible clusterings of the categories of a categorical predictor can be useful, especially for cases where some categories might have too few observations to obtain accurate estimates of their coefficients, or in the presence of user-defined constraints on the shape of these clusters.

\section*{Acknowledgements}{This research has been financed in part by research projects EC H2020 MSCA RISE NeEDS (Grant agreement ID: 822214),  FQM-329 and P18-FR-2369 (Junta de Andaluc\'{\i}a), and PID2019-110886RB-I00 (Ministerio de Ciencia, Innovaci\'on y Universidades, Spain). This support is gratefully acknowledged.}


\begin{thebibliography}{}

\bibitem[Bonhomme and Manresa, 2015]{bonhomme2015grouped}
Bonhomme, S. and Manresa, E. (2015).
\newblock Grouped patterns of heterogeneity in panel data.
\newblock {\em Econometrica}, 83(3):1147--1184.

\bibitem[Carrizosa et~al., 2010]{CarrizosaIJOC10}
Carrizosa, E., Mart\'{\i}n{-}Barrag\'{a}n, B., and Romero~Morales, D. (2010).
\newblock Binarized support vector machines.
\newblock {\em INFORMS Journal on Computing}, 22(1):154--167.

\bibitem[Carrizosa et~al., 2011]{carrizosaEJOR11}
Carrizosa, E., Mart\'{\i}n{-}Barrag\'{a}n, B., and Romero~Morales, D. (2011).
\newblock Detecting relevant variables and interactions in supervised
  classification.
\newblock {\em European Journal of Operational Research}, 213(1):260--269.

\bibitem[Carrizosa et~al., 2017]{carrizosa2017clustering}
Carrizosa, E., Nogales-G{\'o}mez, A., and Romero~Morales, D. (2017).
\newblock Clustering categories in support vector machines.
\newblock {\em Omega}, 66:28--37.

\bibitem[Cerda et~al., 2018]{cerda2018similarity}
Cerda, P., Varoquaux, G., and K{\'e}gl, B. (2018).
\newblock Similarity encoding for learning with dirty categorical variables.
\newblock {\em Machine Learning}, 107(8-10):1477--1494.

\bibitem[Deb and Trivedi, 1997]{deb1997demand}
Deb, P. and Trivedi, P.~K. (1997).
\newblock Demand for medical care by the elderly: a finite mixture approach.
\newblock {\em Journal of Applied Econometrics}, 12(3):313--336.

\bibitem[Detmer et~al., 2020]{detmer2020note}
Detmer, F.~J., Cebral, J., and Slawski, M. (2020).
\newblock A note on coding and standardization of categorical variables in
  (sparse) group lasso regression.
\newblock {\em Journal of Statistical Planning and Inference}.

\bibitem[Dua and Graff, 2017]{Dua:2019}
Dua, D. and Graff, C. (2017).
\newblock {UCI} {M}achine {L}earning {R}epository.
\newblock http://archive.ics.uci.edu/ml.

\bibitem[Hastie et~al., 2015]{hastie2015statistical}
Hastie, T., Tibshirani, R., and Wainwright, M. (2015).
\newblock {\em Statistical Learning with sparsity: The Lasso and
  Generalizations}.
\newblock Chapman and Hall/CRC.

\bibitem[Johannemann et~al., 2019]{johannemann2019sufficient}
Johannemann, J., Hadad, V., Athey, S., and Wager, S. (2019).
\newblock Sufficient representations for categorical variables.
\newblock {\em arXiv preprint arXiv:1908.09874}.

\bibitem[LeBlanc and Tibshirani, 1998]{leblanc1998monotone}
LeBlanc, M. and Tibshirani, R. (1998).
\newblock Monotone shrinkage of trees.
\newblock {\em Journal of Computational and Graphical Statistics},
  7(4):417--433.

\bibitem[Meier et~al., 2008]{meier2008group}
Meier, L., Van De~Geer, S., and B{\"u}hlmann, P. (2008).
\newblock The group lasso for logistic regression.
\newblock {\em Journal of the Royal Statistical Society: Series B (Statistical
  Methodology)}, 70(1):53--71.

\bibitem[Mikolov et~al., 2013]{Mikolov2013EfficientEO}
Mikolov, T., Chen, K., Corrado, G.~S., and Dean, J. (2013).
\newblock Efficient estimation of word representations in vector space.
\newblock {\em International Conference on Learning Representations}, arXiv
  preprint arXiv: 1301.3781.

\bibitem[Moeyersoms et~al., 2016]{moeyersoms2016explaining}
Moeyersoms, J., d'Alessandro, B., Provost, F., and Martens, D. (2016).
\newblock Explaining classification models built on high-dimensional sparse
  data.
\newblock {\em arXiv preprint arXiv:1607.06280}.

\bibitem[Resende and Ribeiro, 2016]{resende2016optimization}
Resende, M.~G. and Ribeiro, C.~C. (2016).
\newblock {\em Optimization by GRASP}.
\newblock Springer.

\bibitem[Van~de Geer, 2008]{van2008high}
Van~de Geer, S.~A. (2008).
\newblock High-dimensional generalized linear models and the lasso.
\newblock {\em The Annals of Statistics}, 36(2):614--645.

\end{thebibliography}
\end{document}